\documentclass{article}

\PassOptionsToPackage{square,numbers}{natbib}

\usepackage[preprint]{neurips_2023}

\usepackage[utf8]{inputenc} 
\usepackage[T1]{fontenc}    
\usepackage{booktabs}       
\usepackage{tabularx} 
\usepackage{amsfonts}       
\usepackage{nicefrac}       
\usepackage{microtype}      
\usepackage{xcolor}         
\usepackage{amsmath}
\usepackage{amssymb} 
\usepackage{bm} 
\usepackage{amsthm} 
\usepackage{xfrac}
\usepackage{multirow}  
\usepackage{enumerate} 
\usepackage{enumitem} 
\usepackage{pgfplots} 
\usepackage{caption} 
\usepackage{subcaption} 
\usepackage[breaklinks=true,bookmarks=false]{hyperref}
\usepackage{url}
\usepackage{siunitx} 
\usepackage{soul}

\hypersetup{
    colorlinks = true,
    urlcolor = blue,
    citecolor=.,
    linkcolor=.,
}
\usepackage{cleveref}
\usepackage{tikz} 
\usetikzlibrary{patterns}
\usetikzlibrary{positioning}
\usetikzlibrary{fadings}

\newcolumntype{Y}{>{\centering\arraybackslash}X}

\usepackage[colorinlistoftodos]{todonotes}

\newcommand{\R}{\mathbb{R}} 

\def\rvepsilon{{\bm{\epsilon}}}

\def\rva{{\mathbf{a}}}

\def\rvc{{\mathbf{c}}}

\def\rvi{{\mathbf{i}}}

\def\rvv{{\mathbf{v}}}

\def\rvx{{\mathbf{x}}}

\def\rvz{{\mathbf{z}}}

\def\vzero{{\bm{0}}}

\def\mI{{\bm{I}}}

\DeclareMathAlphabet{\mathsfit}{\encodingdefault}{\sfdefault}{m}{sl}
\SetMathAlphabet{\mathsfit}{bold}{\encodingdefault}{\sfdefault}{bx}{n}

\title{GAIA-2: A Controllable Multi-View \\ Generative  World Model for Autonomous Driving}

\author{%
  Lloyd Russell\thanks{Equal contribution, random order.}\\
  \And
  Anthony Hu\footnotemark[1]\\
  \And
  Lorenzo Bertoni\footnotemark[1]\\
  \And
  George Fedoseev\footnotemark[1]\\
  \AND
  Jamie Shotton\\
  \And
  Elahe Arani\\
  \And
  Gianluca Corrado\footnotemark[1]\\
  \AND
  Wayve\\
  \texttt{research@wayve.ai}\\
}

\begin{document}
\graphicspath{{Figures/}}

\maketitle

\begin{abstract}
    Generative models offer a scalable and flexible paradigm for simulating complex environments, yet current approaches fall short in addressing the domain-specific requirements of autonomous driving—such as multi-agent interactions, fine-grained control, and multi-camera consistency. We introduce GAIA-2, \textit{Generative AI for Autonomy}, a latent diffusion world model that unifies these capabilities within a single generative framework. GAIA-2 supports controllable video generation conditioned on a rich set of structured inputs: ego-vehicle dynamics, agent configurations, environmental factors, and road semantics. It generates high-resolution, spatiotemporally consistent multi-camera videos across geographically diverse driving environments (UK, US, Germany). The model integrates both structured conditioning and external latent embeddings (e.g., from a proprietary driving model) to facilitate flexible and semantically grounded scene synthesis. Through this integration, GAIA-2 enables scalable simulation of both common and rare driving scenarios, advancing the use of generative world models as a core tool in the development of autonomous systems.
    Videos are available at \url{https://wayve.ai/thinking/gaia-2}.
\end{abstract}

\section{Introduction}
\label{section:introduction}

Realistic simulation of driving scenarios is a foundational requirement for the development, training, and evaluation of autonomous driving systems. Generative world models enable scalable and diverse synthetic data creation, reducing reliance on expensive real-world data collection and facilitating robust evaluation in safe and repeatable environments. Unlike general-purpose text-to-video or image-to-video models, which primarily focus on visual realism and temporal coherence, autonomous driving applications demand fine-grained control over domain-specific aspects of the scene.

Generative models for autonomous driving must accurately simulate factors such as the actions of the ego-vehicle, the locations and movements of other agents (e.g., vehicles, pedestrians, cyclists), and their interactions. Moreover, these models must allow conditional generation based on contextual attributes, such as geographic location, weather, time of day, road configuration (e.g., speed limits, number of lanes, pedestrian crossings, traffic lights, intersections), and rare but critical edge-case scenarios. The ability to generate consistent multi-camera video streams is also essential, as autonomous vehicles rely on spatially and temporally coherent input from multiple perspectives for perception and decision-making. Meeting these domain-specific constraints presents unique challenges that are not adequately addressed by existing video generative models.

While recent advancements in latent video generation, particularly those leveraging continuous~\cite{kingma14,agarwal2025cosmos} or quantized~\cite{agarwal2025cosmos,oord17,esser21} latent spaces, have led to improvements in efficiency and visual quality, current approaches remain limited in scope. In the context of autonomous driving, several models have introduced task-specific conditioning mechanisms~\cite{gaia1-2023,wang2023drivedreamer,wang2024driving,gao2024vista,zhao2024drivedreamer4d}, enabling some control over scenario elements. However, existing models typically support only a subset of the required capabilities, such as limited conditioning types, lack of multi-camera generation, or poor spatial-temporal coherence. As such, a comprehensive and unified generative framework remains an open challenge in the field.

To address this gap, we introduce \textit{GAIA-2}, a domain-specialized latent diffusion model that represents a significant advancement in video-generative world modelling for autonomous driving. GAIA-2 advances prior work by supporting a conditional generation of high-resolution, multi-camera driving scenes, with fine-grained control over ego-vehicle actions, agent behavior, scene geometry, and environmental factors. The model can generate up to five temporally and spatially consistent camera streams at a resolution of $448\times960$, accommodating various multi-camera rig configurations used in real-world autonomous vehicle systems.

\begin{figure}[t]
    \centering
    \includegraphics[width=1.0\textwidth]{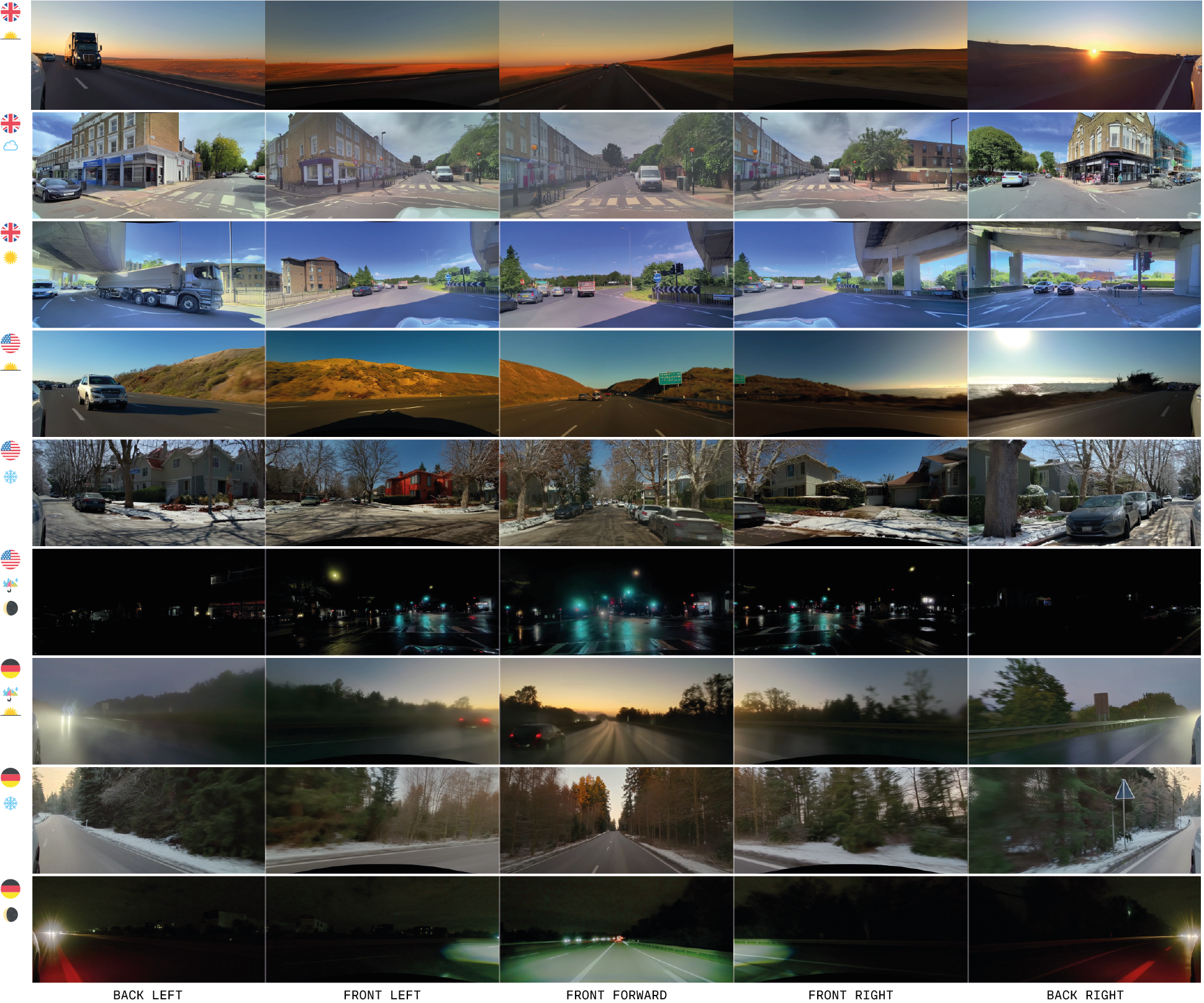}
    \caption{A selection of `\textit{from scratch}' generations demonstrating the diversity of synthetic scenarios possible with GAIA-2.}
    \label{fig:diversity}
    \vspace{-10pt}
\end{figure}

GAIA-2 supports conditioning on a wide range of scene attributes, including ego-vehicle kinematics (e.g., speed, curvature), geographic regions (UK, US, Germany), time of day, weather, and a rich taxonomy of road layout features—such as number and type of lanes (e.g., drivable, bus, cycle), presence of pedestrian crossings, traffic lights, and intersection topology (e.g., one-way roads, roundabouts). It also allows direct control over the locations, orientations, and dimensions of dynamic agents within the scene. This extensive set of conditioning parameters allows for precise control over the generated scenarios, enabling simulation of both typical driving conditions and edge cases critical for robust system evaluation.

To ensure flexibility and interoperability, GAIA-2 enables conditioning on external latent spaces, including CLIP embeddings and a proprietary model trained to produce driving-specific latent representations. This capability allows semantic control over scene content and facilitates integration with downstream planning or perception modules. GAIA-2 further supports multiple video generation modes, including generation from scratch, prediction from past context, and selective content editing through inpainting.

By integrating these capabilities into a single framework, GAIA-2 sets a new benchmark in video-generative world models for autonomous driving. It enables detailed, controllable, and realistic simulation across diverse conditions, providing a powerful tool for scalable training and robust evaluation of autonomous driving systems.

\begin{figure}[t]
  \centering
  \makebox[\textwidth][c]{%
    \includegraphics[width=1\textwidth]{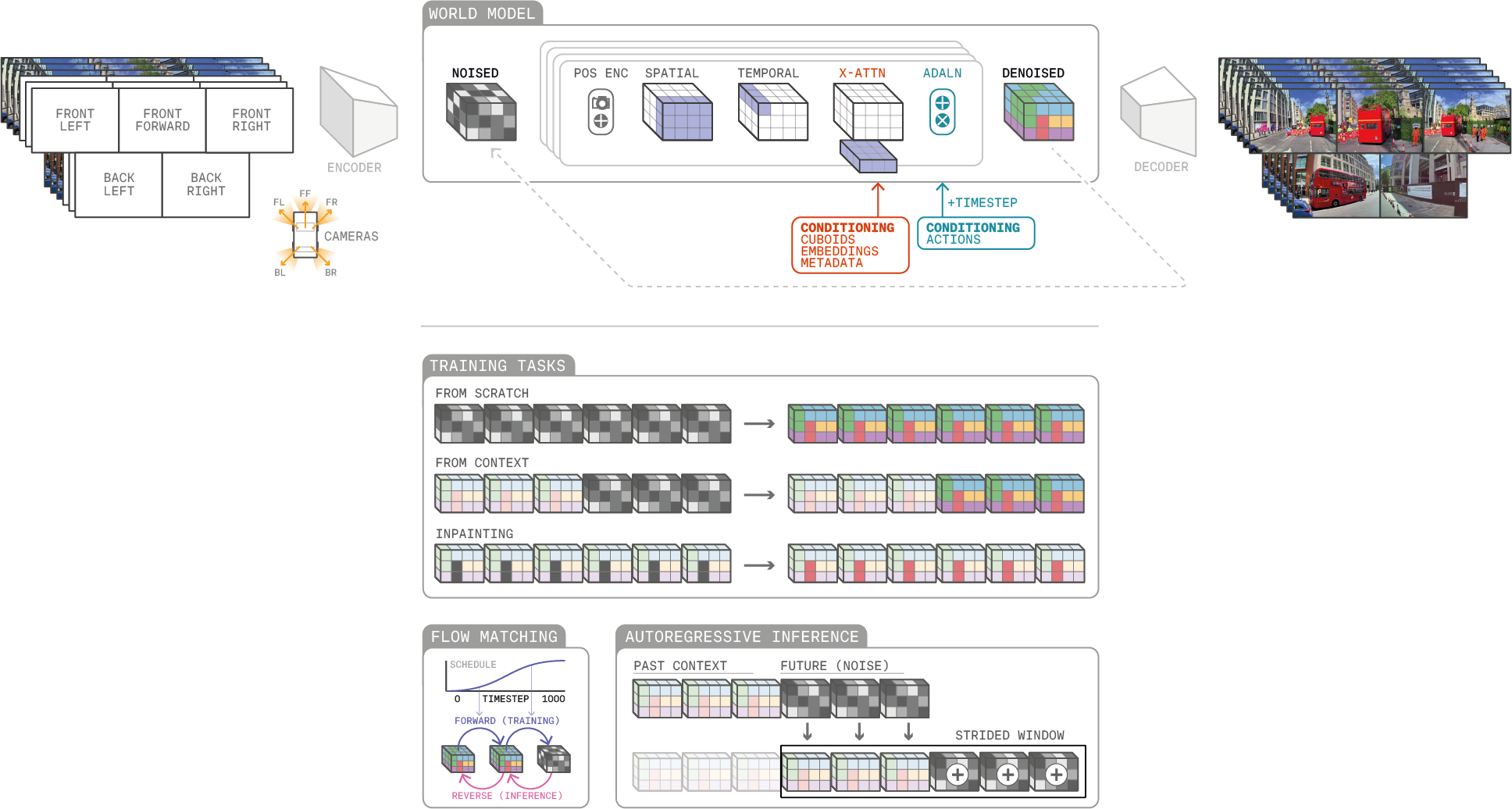}%
  }
  \caption{\textbf{GAIA-2 world model}. The architecture schematic shows full surround camera views independently encoded by a video tokenizer. The combined multi-camera latents are used as past context and linearly interpolated with sampled noise. We add camera parameters along with positional encodings before passing the latents through a space-time factorized transformer. Conditioning is implemented via cross-attention and adaptive layer norm. The latents are denoised into future latents, conditioned on various inputs, including actions, 3D bounding boxes, metadata, and scenario embeddings. The denoised latents are then decoded back to pixel space by the video tokenizer. Below, we depict the various tasks GAIA-2 is trained on, including from scratch generations, 
  context rollouts, and spatial inpainting. At inference, we can generate beyond the window duration GAIA-2 was trained on by taking overlapping strides to generate new frames conditioned on previously generated frames.}
  \label{fig:wm-arch}
  \vspace{-10pt}
\end{figure}

\section{Model}
\label{section:model}
GAIA-2 is a surround-view video generative world model with structured conditioning, multi-camera coherence, and high spatial-temporal resolution. The architecture (\Cref{fig:wm-arch}) is composed of two primary components: a video tokenizer and a latent world model. Together, these modules enable GAIA-2 to generate realistic and semantically coherent video across multiple viewpoints with rich conditional control.

The video tokenizer compresses the raw high-resolution video into a compact, continuous latent space, preserving semantic and temporal structure. This compact representation enables efficient learning and generation at scale. The world model then learns to predict future latent states, conditioned on past latent states, actions, and a diverse set of domain-specific control signals. It can also be used to generate novel states completely from scratch and modify video content through inpainting. The predicted latent states are subsequently decoded back into pixel space using the video tokenizer decoder.

In contrast to most latent diffusion models, GAIA-2 employs a much higher spatial compression rate (e.g., $32\times$ vs. the more typical $8\times$), which is offset by increasing the channel dimension of the latent space (e.g., 64 channels instead of 16). This yields fewer but semantically richer latent tokens. The resulting advantages are twofold: (1) shorter latent sequences enable faster inference and better memory efficiency, and (2) the model demonstrates improved ability to capture video content and temporal dynamics. This parameterization strategy is inspired by prior work on compact latent representations~\cite{dcae,ltx}.

Unlike its predecessor GAIA-1~\cite{gaia1-2023}, which relied on discrete latent variables, GAIA-2 employs a continuous latent space, improving temporal smoothness and reconstruction fidelity. Furthermore, GAIA-2 introduces a flexible conditioning interface that supports ego-vehicle actions, dynamic agent states (e.g., 3D bounding boxes), structured metadata, CLIP and scenario embeddings, and camera geometry. This design enables robust control over scene semantics and agent behavior during generation, while ensuring cross-view consistency and temporal coherence.

\subsection{Video Tokenizer}

The video tokenizer compresses pixel-space video into a compact latent space that is continuous and semantically structured. It is composed of a space-time factorized transformer with an asymmetric encoder-decoder architecture (with 85M and 200M parameters, respectively). The encoder extracts spatiotemporally downsampled latents that are temporally independent; the decoder reconstructs full-frame video from these latents using temporal context for temporal consistency. 

\begin{figure}[t]
  \centering
  \makebox[\textwidth][c]{%
    \includegraphics[width=1\textwidth]{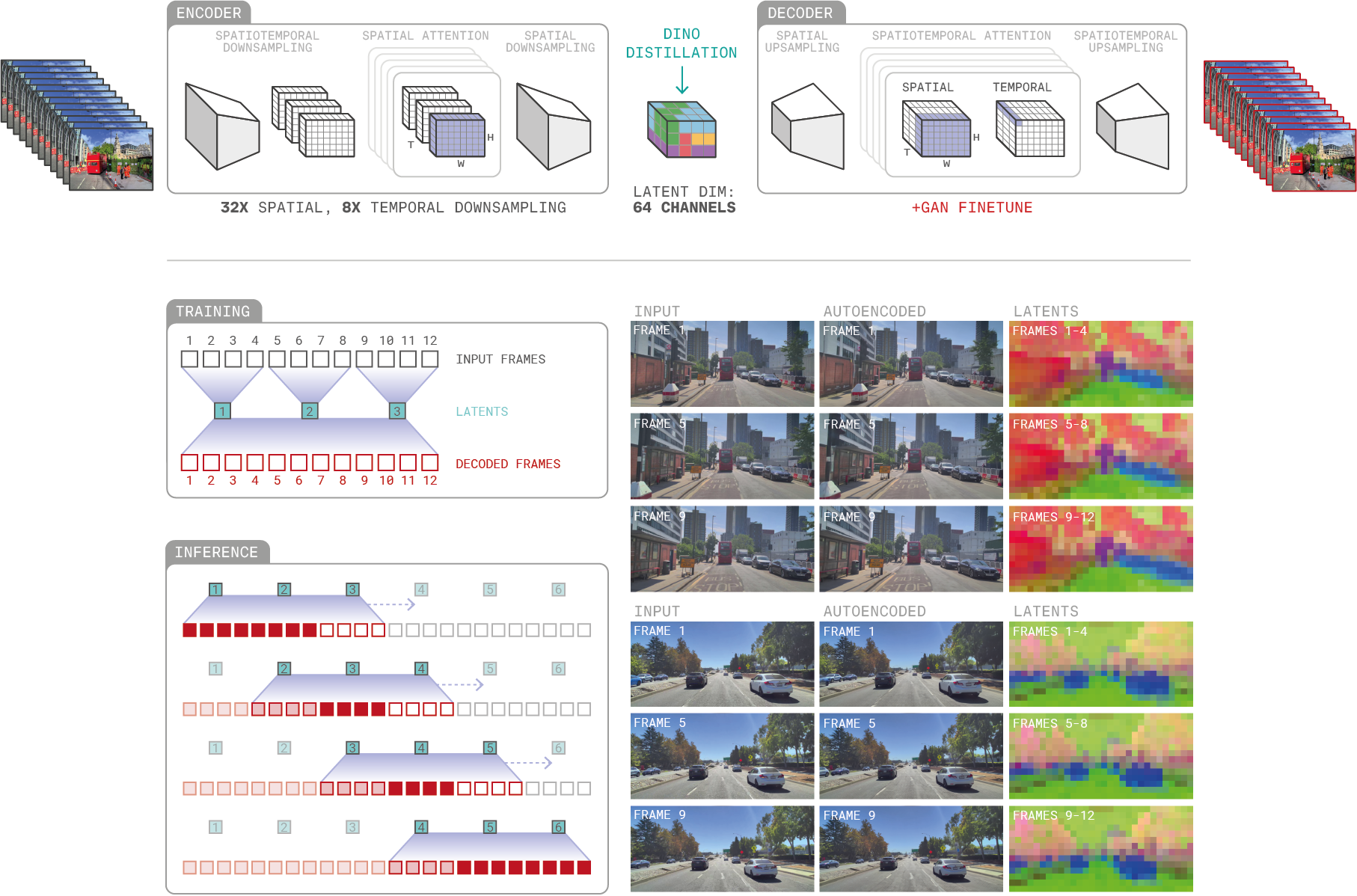}%
  }
  \caption{\textbf{GAIA-2 video tokenizer}. The architecture schematic depicts input frames spatiotemporally downsampled into temporally independent latents. A high spatial compression rate is compensated for by an increased latent channel dimension. Latents are decoded back to video frames, leveraging full spatiotemporal attention to ensure temporal consistency. To autoencode long sequences, we employ a rolling inference mechanism where current latents are decoded using past and future context in a sliding window fashion. The examples show input frames, decoded frames, and a visualization of the latent space. The first 3 principal components of the latent features are mapped to RGB values. Notably, the latent space is semantically stable across frames and across samples.}
    \label{fig:video-tokenizer}
\end{figure}

\subsubsection{Encoder}
Given an input video $(\rvi_1, ..., \rvi_{T_v})$, the encoder $e_{\phi}$ computes latent tokens $(\rvz_1, ..., \rvz_{T_L}) = e_{\phi}(\rvi_1, ..., \rvi_{T_v})$, where $T_v$ corresponds to the number of video frames and $T_L$ the number of latents. Let us denote by $H_v \times W_v$ the spatial resolution of the video frames and $H\times W$ the spatial resolution of the latent tokens. The encoder downsamples spatially by a factor $\frac{H_v}{H} = 32$ and temporally by a factor $\frac{T_v}{T_L} = 8$. The latent dimension is $L=64$, which results in a total compression of ($\frac{T_v \times H_v \times W_v \times 3}{T_L \times H \times W \times L} =384)$ with $(T_v, H_v, W_v) = (24, 448, 960)$ and $(T_L, H, W) = (3, 14, 30)$.

This downsampling is achieved by temporally striding the input frames by $2\times$ and the following modules:
\begin{enumerate}
    \item A downsampling convolutional block with stride $2 \times 8 \times 8$ (time, height, width) followed by another downsampling convolutional block with stride $2 \times 2 \times 2$ (both operating at embedding dimension $512$).
    \item A series of $24$ spatial transformer blocks with dimension $512$ and $16$ heads.
    \item A final convolution with stride $1 \times 2 \times 2$, followed by a linear projection to $2L$ channels to model a Gaussian distribution over latents. Note that the latent dimension is doubled as the encoder predicts the mean and standard deviation of a Gaussian distribution. During both training and inference, the latents are sampled from this resulting distribution.
\end{enumerate}

\subsubsection{Decoder}
The decoder architecture is:
\begin{enumerate}
    \item A linear projection from the latent dimension to the embedding dimension followed by a first upsampling convolutional block with stride $1 \times 2 \times 2$ (upsampling is achieved with a depth-to-space module \citep{shi16}).
    \item A series of 16 space-time factorized transformer blocks with dimension $512$ and $16$ heads.
    \item An upsampling convolutional block with stride $2 \times 2 \times 2$ followed by $8$ space-time factorized transformer blocks with dimension $512$ and $16$ heads.
    \item A final upsampling convolutional block with stride $2 \times 8 \times 8$ and dimension $3$ corresponding to the pixel RGB channels.
\end{enumerate}

A key difference between the encoder and decoder is that the encoder independently maps $8$ consecutive video frames to a single temporal latent, while the decoder jointly decodes the $T_L=3$ temporal latents to $T_v=24$ video frames for temporal consistency. During inference, the video frames are decoded with a sliding window. The logic is shown in diagrams `\textit{Training}' and `\textit{Inference}' of \Cref{fig:video-tokenizer}.

\subsubsection{Training Losses}
The video tokenizer is trained with a combination of pixel reconstruction and latent space losses:
\begin{itemize}
    \item Image reconstruction with $L_1$, $L_2$, and perceptual losses \citep{Johnson2016Perceptual}.
    \item DINO \citep{oquab2023dinov2} distillation on the latent features through a cosine similarity loss, encouraging semantic alignment with pre-trained representations. The video frames are encoded with the DINO model and temporally downsampled with linear interpolation to match the dimensionality of the latent features.
    \item Kullback-Leibler divergence loss \citep{kingma14} with respect to a standard Gaussian distribution to regularize the latent space. 
\end{itemize}

To improve visual quality, we further fine-tune the decoder with a GAN loss \citep{esser21} using a 3D convolutional discriminator and the image reconstruction losses, while keeping the encoder frozen. The discriminator is a series of residual 3D convolutional blocks with base channel 64, stride 2 in time and space with 3D blur pooling \citep{zhang2019shiftinvar}, channel multipliers [2, 4, 8, 8], 3D instance normalization, and LeakyReLU with slope $0.2$. It uses spectral normalization \citep{miyato18} and the original GAN loss implemented with a softplus activation function.

\subsection{World Model}
The latent world model predicts future latent states conditioned on past latents, actions, and a rich set of conditioning inputs. It is implemented as a space-time factorized transformer with 8.4B parameters and is trained using flow matching~\cite{lipman23} for stability and sample efficiency.

Let $\rvx_{1:T} \in \mathbb{R}^{T \times N \times H \times W \times L}$ be the input latents, where $T$ is the temporal window and $N$ is the number of cameras. The input latents are obtained by independently encoding each camera view with the encoder $e_{\phi}$. At each timestep $t$, we also provide an action vector $\rva_t$ and a conditioning vector $\rvc_t$.

\subsubsection{Architecture}

The world model is a space-time factorized transformer with hidden dimension $C$. Each action $\rva_t$ is embedded to $\R^C$, and the conditioning vector $\rvc_t$ embedded to $\R^{K\times C}$, where $K$ corresponds to the number of conditioning variables. The flow matching time $\tau \in [0, 1]$ is also mapped to $\R^C$ using a sinusoidal encoding \cite{ho2020}.

Flow matching time $\tau$ and action $\rva_t$ are injected into each transformer block through the adaptive layer norm \cite{peebles2023scalable}, while we use cross-attention for the other conditioning variables $\rvc_t$. We have found that action conditioning was more accurate when using adaptive layer norm over cross-attention. As the action affects every spatial token, the adaptive layer norm provides an explicit information gateway instead of having to rely on a learnable attention mechanism.

Regarding positional encoding, we independently encode: (i) spatial token position with a sinusoidal embedding, (ii) camera timestamp with a sinusoidal embedding followed by a small MLP, and (iii) camera geometry (distortion, intrinsic, and extrinsics) with learnable linear layers. All these positional encodings are added to the input latents at the beginning of each transformer block, similar to \cite{polyak2024movie}. 

The world model contains $22$ space-time factorized transformer blocks with hidden dimension $C=4096$ and $32$ heads. Each transformer block contains a spatial attention (over space and cameras), a temporal attention, a cross-attention, and an MLP layer with an adaptive layer norm. For increased training stability, we use query-key normalization \cite{dehghani2023scaling} before each attention layer.

\subsubsection{Losses}
At training time, we randomly sample the number of context frames $t \in \{0, ..., T-1\}$ where $t=0$ corresponds to from scratch generation. We also sample a flow matching time $\tau \in [0, 1]$ according to a pre-defined distribution (see \Cref{subsubsection:flow-matching-distribution} for more details). The context latents $\rvx_{1:t} $ are unchanged, while the future latents $\rvx_{t+1:T} $ are linearly interpolated with random Gaussian noise $\rvepsilon_{t+1:T} \sim \mathcal{N}(\vzero, \mI)$. 
\begin{equation}
    \rvx_{t+1:T}^{\tau} = \tau \rvx_{t+1:T} + (1 - \tau) \rvepsilon_{t+1:T}
\end{equation}

The velocity target vector $\rvv_{t+1:T}$ is the difference between target latents and random noise:
\begin{equation}
    \rvv_{t+1:T} = \rvx_{t+1:T} - \rvepsilon_{t+1:T}
\end{equation}

The world model $f_{\theta}$ predicts the target velocity $\rvv_{t+1:T}$ conditioned on context latents $\rvx_{1:t}$, actions $\rva_{1:T}$ and conditioning variables $\rvc_{1:T}$.
\begin{equation}
    \hat{\rvv}_{t+1:T} = f_{\theta}(\rvx_{t+1:T}^{\tau} | \rvx_{1:t}, \rva_{1:T}, \rvc_{1:T}, \tau)
\end{equation}

The model is trained with an $L_2$ loss between the predicted and the target velocity.
\begin{equation}
    \mathcal{L}_{\text{world-model}} = \mathbb{E}_{t \sim P(t), \tau \sim p(\tau)} [||\rvv_{t+1:T} - \hat{\rvv}_{t+1:T} ||^2]
\end{equation}
where $t \sim P(t)$ denotes the distribution of context frames and $\tau \sim p(\tau)$ the distribution of flow matching time.

\subsubsection{Conditioning}
\label{subsubsection:conditioning}
GAIA-2 supports rich and structured conditioning inputs that enable fine-grained control over the generated scenes. These inputs include ego-vehicle actions, dynamic agent properties, scene-level metadata, camera configurations, timestamp embedding, and external latent representations such as CLIP or proprietary scenario embeddings. The conditioning mechanisms are integrated into the world model through a combination of adaptive layer normalization (for action), additive modules (for camera geometry and timestamp), and cross-attention (for all other variables).

\paragraph{Camera Parameters.} We compute separate embeddings for intrinsics, extrinsics, and distortion, which are then summed to form a unified camera encoding. For intrinsics, we extract focal lengths and principal point coordinates from the intrinsic matrix, normalize them, and project them into a shared latent space. Extrinsics and distortion coefficients are similarly processed through their respective encoders to yield compact representations. This configuration allows the model to effectively incorporate real-world camera variations. \Cref{fig:camera-rigs} illustrates the top three most common camera rig configurations in the training dataset.

\paragraph{Video Frequency.} To account for variable video frame rates, GAIA-2 uses timestamp conditioning. Each timestamp is: (\textit{i}) Normalized relative to the present time and scaled to the range $[-1, 1]$, (\textit{ii}) Transformed using sinusoidal functions (Fourier feature encoding), and (\textit{iii}) Passed through an MLP to produce a vector in the shared latent space. This encoding captures both low- and high-frequency temporal variation and enables the model to reason effectively over videos recorded at different rates.

\paragraph{Action.} The ego-vehicle action is parameterized by speed and curvature. Since these quantities span multiple orders of magnitude, we use a symmetric logarithmic transformation \textit{symlog} \cite{webber12} for normalization:
\begin{equation}
    \text{symlog}(y) = \text{sign}(y)
    \frac{\log(1 + s|y|)}{\log(1 + s|y_{\text{max}}|)}
\end{equation}
Here, $y$ represents speed or curvature, and $s$ is a scale factor:
For curvature (units: $m^{-1}$, range: $0.0001$ to $0.1$), we use $s = 1000$ to amplify low values. For speed (units: m/s, range: $0$ to $75$), we use $s = 3.6$ to express values in km/h.
The result is a compact representation scaled to $[-1, 1]$, improving training stability.

\paragraph{Dynamic Agents.}
To represent surrounding agents, we use 3D bounding boxes predicted by a 3D object detector~\cite{wang2023streampetr} re-trained on our dataset. Each box encodes the 3D location, orientation, dimensions, and category of an agent. The 3D boxes are projected into the 2D image plane and normalized, yielding $f_i \in \mathbb{R}^{T \times N \times B \times 13}$ conditioning features where $T$ denotes the number of temporal latents, $N$ the number of cameras, and $B$ the maximum number of 3D bounding boxes (zero-padding as needed). Each feature dimension is embedded independently and aggregated via a single-layer MLP.

To enhance model robustness and generalizability, we implement dropout at both the feature dimension and instance levels during training. Specifically, feature dimensions are dropped out with a probability of $p = 0.3$, allowing the model to operate under incomplete information at inference. This setup allows, for example, conditioning on 2D projected boxes without specifying the 3D locations of instances, or omitting orientations to let the model predict the most plausible orientation based on other conditions.

 At the instance level, for each camera, we sample a frame $t \in \{1, ..., T\}$ and calculate the number of detected instances $N_{\text{instances}}$. We then sample the number of instances to condition on $n\in \{0, ..., \min(B, N_{\text{instances}})\}$, and apply dropout to the instances exceeding this sample size. This allows the model to adapt to a variable number of dynamic agents during inference. Note that while keeping $n$ constant across time, we do not use instance tracking, allowing the model to independently determine whether the conditioning features across frames belong to the same or different instances.

\paragraph{Metadata.} Metadata features are categorical and embedded using dedicated learnable embedding layers. These include: Country, weather, time of day; Speed limits; Number and types of lanes (e.g., bus, cycle); Pedestrian crossings, traffic lights and their states; One-way road indicators and intersection types.
These embeddings allow GAIA-2 to learn nuanced relationships between scene-level features and their effects on behavior, enabling simulation of both typical and rare scenarios.

\paragraph{CLIP Embedding.} To enable semantic scene conditioning, GAIA-2 supports conditioning on CLIP embeddings~\cite{radford21}. During training, we extract CLIP features using the image encoder on video frames. At inference, these can be replaced with CLIP text encoder outputs from natural language prompts. All CLIP embeddings are projected into the model’s latent space using a learnable linear projection.
This enables zero-shot control over scene semantics through natural language or visual similarity.

\paragraph{Scenario Embedding.} GAIA-2 can also be conditioned on scenario embeddings obtained from an internal proprietary model trained to encode driving-specific information. These embeddings compactly capture ego-action and scene context, such as road layout and agent configurations. The scenario vectors are projected via a learnable linear layer into the latent space before integration into the transformer.
This allows high-level scenario generation from a compact abstract representation.

\subsubsection{Flow matching Time Distribution}
\label{subsubsection:flow-matching-distribution}
A critical factor for training the world model under the flow matching framework is the distribution of the flow matching time $\tau$. This distribution determines how frequently the model sees latent inputs that are close to real latents versus heavily perturbed.

We use a bimodal logit-normal distribution with two modes:
\begin{itemize}
    \item A primary mode centered at $\mu = 0.5$, $\sigma = 1.4$, sampled with probability $p = 0.8$. This biases the model towards learning with low-to-moderate noise levels. Empirically, this encourages learning useful gradients, as even small amounts of noise can significantly perturb high-capacity latents.
    \item A secondary mode centered at $\mu = -3.0$, $\sigma = 1.0$, sampled with probability $p = 0.2$. This concentrates training on nearly pure noise regions around $\tau = 0$, helping the model learn spatial structures and low-level dynamics, such as ego-motion or object trajectories.
\end{itemize}
This bimodal strategy ensures that training is effective across both low and high-noise regimes, improving generalization and sample quality.

In addition, the input latents $\rvx_t$ are normalized by their mean $\mu_x$ and standard deviation $\sigma_x$, following~\cite{rombach2022highresolution}, to ensure their magnitude matches that of the added Gaussian noise. This avoids scale mismatch between the signal and the perturbation, which can otherwise degrade training dynamics.
\section{Data}
\label{section:data}

GAIA-2 is trained on a large-scale internal dataset specifically curated to support the diverse demands of video generation for autonomous driving. The dataset comprises approximately 25 million video sequences, each spanning 2 seconds, collected between 2019 and 2024. Recordings were obtained across three countries—the United Kingdom, the United States, and Germany—to ensure coverage of geographically and environmentally diverse driving conditions.

To capture the complexity of real-world autonomous driving, data collection involved multiple vehicle platforms, including three different car models and two van types. Each vehicle was outfitted with either five or six cameras, configured to provide comprehensive 360-degree surround-view coverage. The camera systems varied in their capture frequencies—20 Hz, 25 Hz, and 30 Hz—introducing a range of temporal resolutions. This variability reflects the heterogeneous nature of sensor setups deployed in actual autonomous vehicles and supports GAIA-2’s ability to generalize across different input rates and hardware specifications.

An important characteristic of the dataset is the variability in camera placement throughout the data collection period. Over time, the positions and calibrations of the cameras were adjusted across platforms, introducing a broad spectrum of spatial configurations. This diversity provides a strong training signal for generalizing across different camera rigs, a key requirement for scalable synthetic data generation in the autonomous driving domain.

The dataset also encompasses a wide range of driving scenarios, including diverse weather conditions, times of day, road types, and traffic environments. To ensure coverage across this complexity, we explicitly balance the training data not just on individual features, but on their joint probability distribution. This approach enables a more representative learning signal by modeling realistic co-occurrences—e.g., specific lighting and weather conditions in certain geographies or behaviors unique to particular road types. To prevent redundant training samples, we enforce a minimum temporal stride between selected sequences, reducing the risk of duplication while maintaining a natural distribution.

For evaluation, we implement a geographically held-out validation strategy. Specific validation geofences are defined to exclude certain regions from the training set entirely. This ensures that model evaluation is performed on unseen locations, providing a more rigorous assessment of generalization performance across different environments.

In summary, this dataset provides a robust foundation for training GAIA-2. Its extensive temporal and spatial coverage, diversity in vehicle and camera configurations, and principled validation setup make it uniquely well-suited for developing generative world models capable of producing realistic and controllable driving video across varied real-world conditions.
\section{Training Procedure}
\label{section:training-procedure}
This section describes the training procedures for both components of GAIA-2: the video tokenizer and the world model. Each component is trained independently using large-scale compute infrastructure and tailored loss configurations to optimize their respective objectives.

\paragraph{Video Tokenizer.} The video tokenizer was trained for $300{\scriptstyle,}000$ steps with a batch size of $128$ using 128 H100 GPUs. Input sequences consisted of $T_v=24$ video frames sampled at their native capture frequencies (20, 25, or 30 Hz). Random spatial crops of size $448 \times 960$ were extracted from the frames. For each training sample, a camera view was randomly selected from the available $N = 5$ perspectives. Notably, each camera stream was encoded independently.

The tokenizer performs $8\times$ temporal and $32\times$ spatial downsampling, yielding a compressed representation with latent dimension $L = 64$, and an effective total compression rate of approximately 400 ($\frac{24 \times 448 \times 960 \times 3}{3 \times 14 \times 30 \times 64} \simeq 400$).

The tokenizer’s loss function is composed of a combination of image reconstruction, perceptual, and semantic alignment terms: (1) DINO v2 (Large)~\cite{oquab2023dinov2} distillation in the latent space with a weight of $0.1$. (2) KL divergence~\cite{kingma14} between the latent distribution and a unit Gaussian, with a low weight of $1\mathrm{e}{-6}$ to encourage smoothness. (3) Pixel-level losses: $L_1$ loss (weight $0.2$), $L_2$ loss (weight $2.0$), and LPIPS perceptual loss~\cite{zhang2018unreasonable} (weight $0.1$).

An exponential moving average (EMA) of the tokenizer parameters $\phi$ was maintained throughout training, with a decay factor of $0.9999$ and updates at every training step. The EMA weights were used at inference.

Training was optimized using AdamW with: $2{\scriptstyle,}500$ warm-up steps to the base learning rate of $1\mathrm{e}{-4}$, and $5{\scriptstyle,}000$ cooldown steps to a final learning rate of $1\mathrm{e}{-5}$; Adam betas $[0.9, 0.95]$, weight decay $0.1$, and gradient clipping at $1.0$. After initial training, the tokenizer decoder was fine-tuned for an additional $20{\scriptstyle,}000$ steps using a GAN loss (weight $0.1$) in combination with the previous reconstruction losses. The discriminator was optimized with a learning rate of $1\mathrm{e}{-5}$.

\paragraph{World Model.} The latent world model was trained for $460{\scriptstyle,}000$ steps with a batch size of $256$ on 256 H100 GPUs. Inputs consisted of $48$ video frames at native capture frequencies (20, 25, or 30 Hz), spatial resolution $448 \times 960$ and across $N=5$ cameras. After encoding these videos to the latent space, this corresponds to $T\times N \times H \times W = 6 \times 5 \times 14 \times 30 = 12{\scriptstyle,}600$ input tokens.

To encourage generalization, we sampled different training tasks as follows:
70\% from-scratch generation, 20\% contextual prediction, and 10\% spatial inpainting. To regularize the model and enable classifier-free guidance, conditioning variables were randomly dropped. Each individual conditioning variable was independently dropped with 80\% probability, and all of them were simultaneously dropped with 10\% probability. 

Input camera views were randomly dropped with 10\% probability to enhance robustness to partial observability.
Latent tokens were normalized using a fixed mean of $\mu_x = 0.0$ and standard deviation $\sigma_x = 0.32$, as empirically determined during tokenizer training.

As with the tokenizer, we maintained an EMA of the world model parameters $\theta$ with a decay factor of $0.9999$ and updates at every training step. EMA weights were used for inference. The optimizer was AdamW with: $2{\scriptstyle,}500$ warm-up steps to an initial learning rate of $5\mathrm{e}{-5}$ and cosine decay over the full training duration to a final learning rate of $6.5\mathrm{e}{-6}$; Adam betas $[0.9, 0.99]$, weight decay $0.1$, and gradient clipping $1.0$.

\section{Inference}
\label{section:inference}
The GAIA-2 model supports a range of inference tasks that showcase its flexibility and controllability across video generation scenarios. These tasks are unified by a shared denoising process operating in the latent space, followed by decoding into pixel space via the video tokenizer.

\begin{figure}[t]
    \centering
    \includegraphics[width=1.0\textwidth]{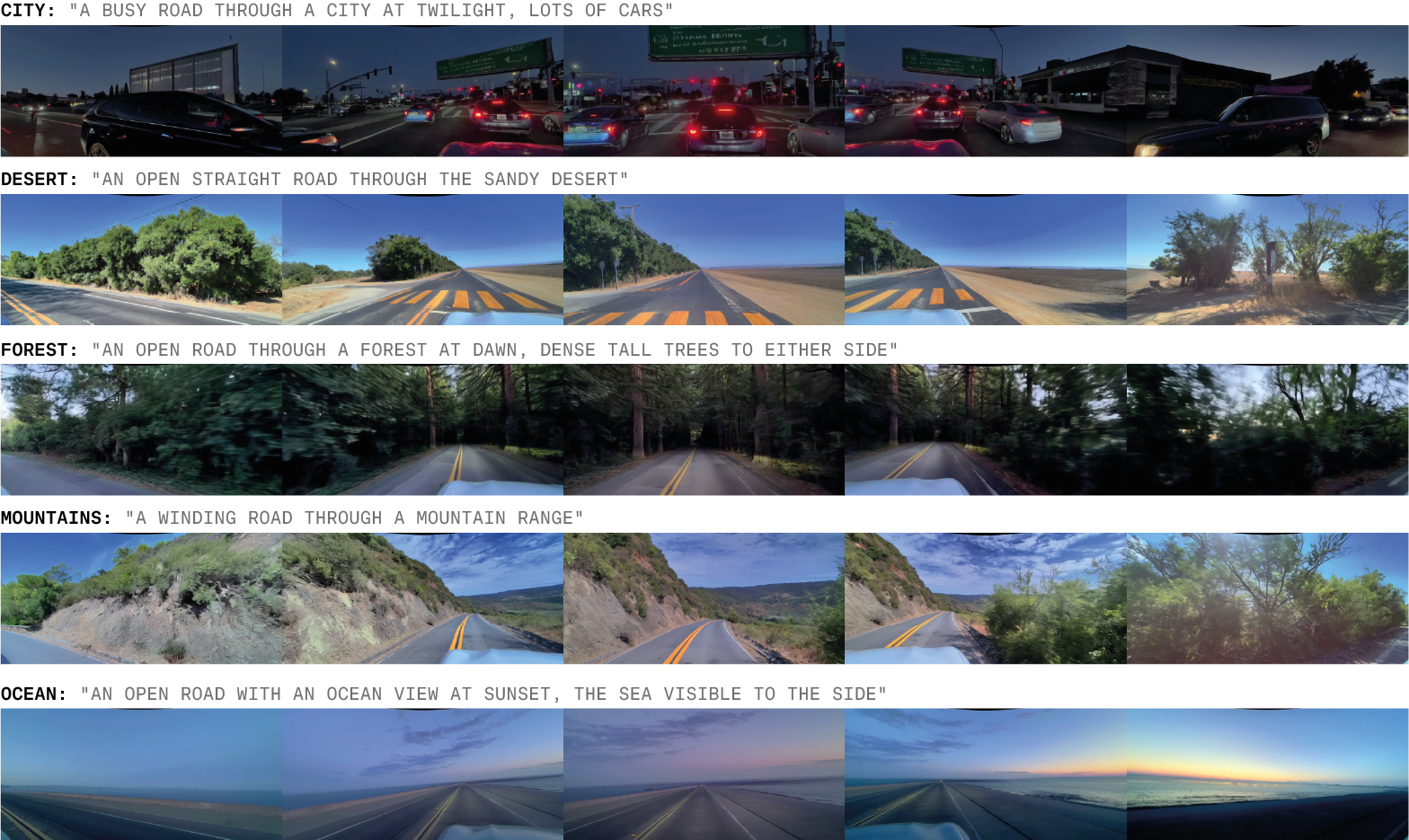}
    \caption{\textbf{CLIP conditioning.} Conditioning GAIA-2 generations with text prompts encoded through the CLIP text encoder provides nuanced control over environmental features.}
    \label{fig:clip}
\end{figure}

\begin{figure}[t]
  \centering
  \includegraphics[width=1.0\textwidth]{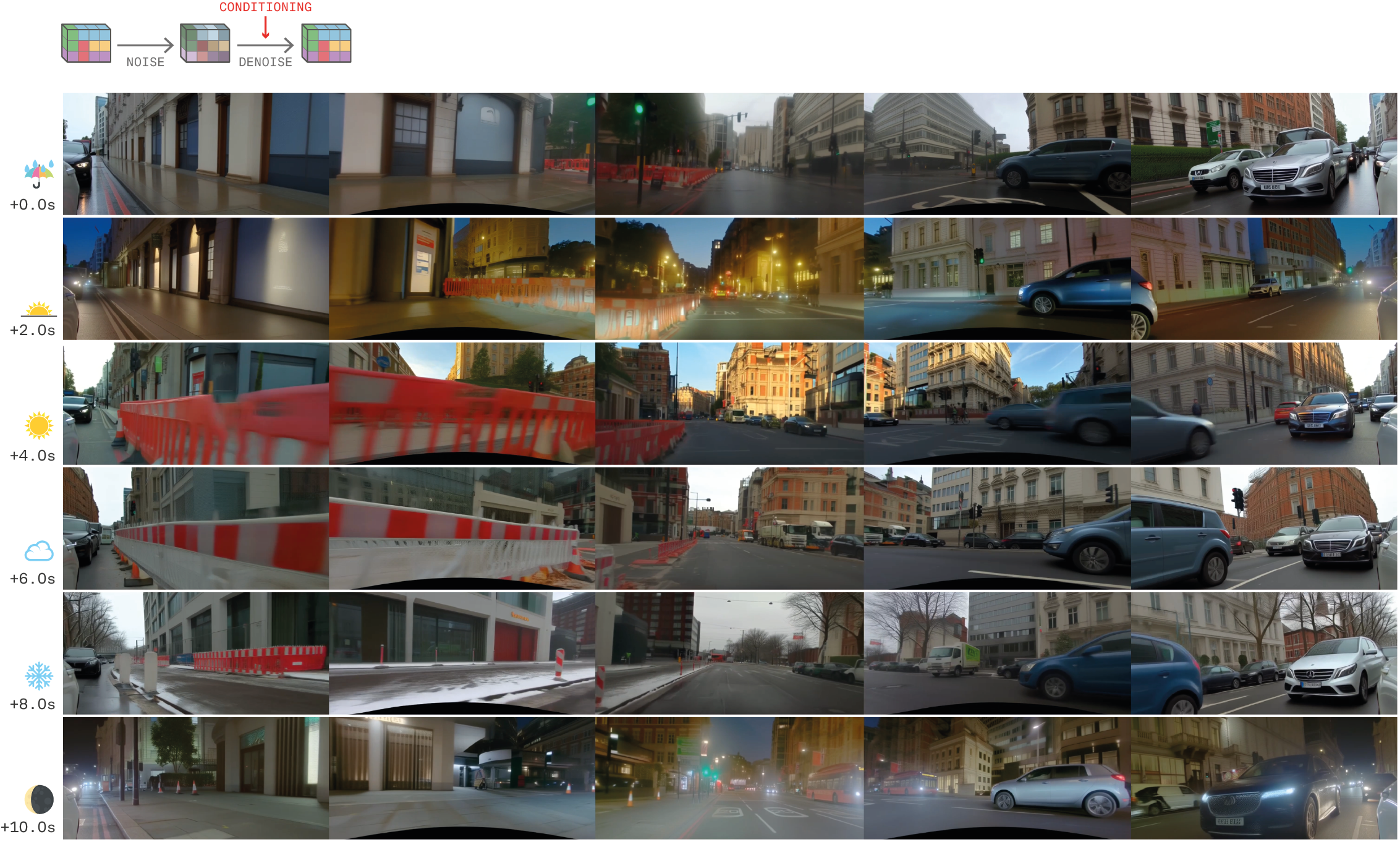}
  \caption{\textbf{Augmentation through partial noising.} By partially noising and denoising video frames, GAIA-2 transforms real video into diverse versions under different environmental settings, such as weather and time of day, while preserving semantic content and ego-actions.}
  \label{fig:noisedenoise}
\end{figure}

\paragraph{Inference Tasks.}
We consider four primary inference modes, each illustrating a distinct capability of the model:

\begin{itemize} 
\item \textbf{Generation from scratch} involves sampling pure Gaussian noise and denoising them with guidance from conditioning variables defined in Section~\ref{subsubsection:conditioning}. The resulting latents are then decoded to video frames using the video tokenizer decoder, with temporally consistent outputs produced via the rolling window decoding mechanism illustrated in Figure~\ref{fig:video-tokenizer}.

\item \textbf{Autoregressive prediction} enables forecasting future latents from a sequence of past context latents. Given an initial context window of $k=3$ temporal latents, the model predicts the next set of latents, appends it to the context, and repeats the process using a sliding window. This approach allows for long-horizon rollouts while incorporating conditioning signals such as ego motion. An example is provided in Figure~\ref{fig:unsafe-actions}.

\item \textbf{Inpainting} allows selective modification of video content. A spatial-temporal mask is applied to the latent input, and the masked regions are regenerated via conditional denoising. Optional guidance from dynamic agent conditioning (e.g., agent locations) can steer the generation within the masked area. An example is shown in~\Cref{fig:inpainting}.

\item \textbf{Scene editing} is achieved by partially noising latents extracted from real video, followed by denoising with altered conditioning. This enables targeted semantic or stylistic transformations—such as changing the weather, time of day, or road layout—without re-generating the full scene. \Cref{fig:noisedenoise} illustrates this capability.
\end{itemize}
These modes demonstrate that GAIA-2 can serve as a general-purpose simulator for a wide variety of scene manipulation tasks, whether starting from noise, context, or existing video.

\paragraph{Inference Noise Schedule.}
For all inference tasks, we adopt the linear-quadratic noise schedule introduced by \cite{polyak2024movie}. This schedule begins with linearly spaced noise levels, which are effective for capturing coarse scene layouts and motion patterns. In later stages, the schedule transitions to quadratically spaced steps that allow more efficient refinement of high-frequency visual details. This hybrid approach improves both generation quality and computational efficiency. In our experiments, we use a fixed number of 50 denoising steps.

\paragraph{Classifier-Free Guidance.}
Classifier-free guidance (CFG) is not used by default during inference. However, for challenging or out-of-distribution scenarios, such as those involving rare edge cases or unusual agent configurations (e.g., Figure~\ref{fig:cuboids}), we activate CFG with a guidance scale ranging from 2 to 20, depending on the complexity of the scene.

In scenarios involving dynamic agent conditioning, where the latent tokens associated with agent-specific regions are known a priori, we apply spatially selective CFG. In this case, guidance is applied only to the spatial locations influenced by the conditioning (e.g., 3D bounding boxes), which enhances generation quality in targeted regions without unnecessarily affecting the rest of the scene. This targeted approach enables more precise control over scene elements while preserving global coherence.

\begin{figure}[t]
    \centering
    \includegraphics[width=1.0\textwidth]{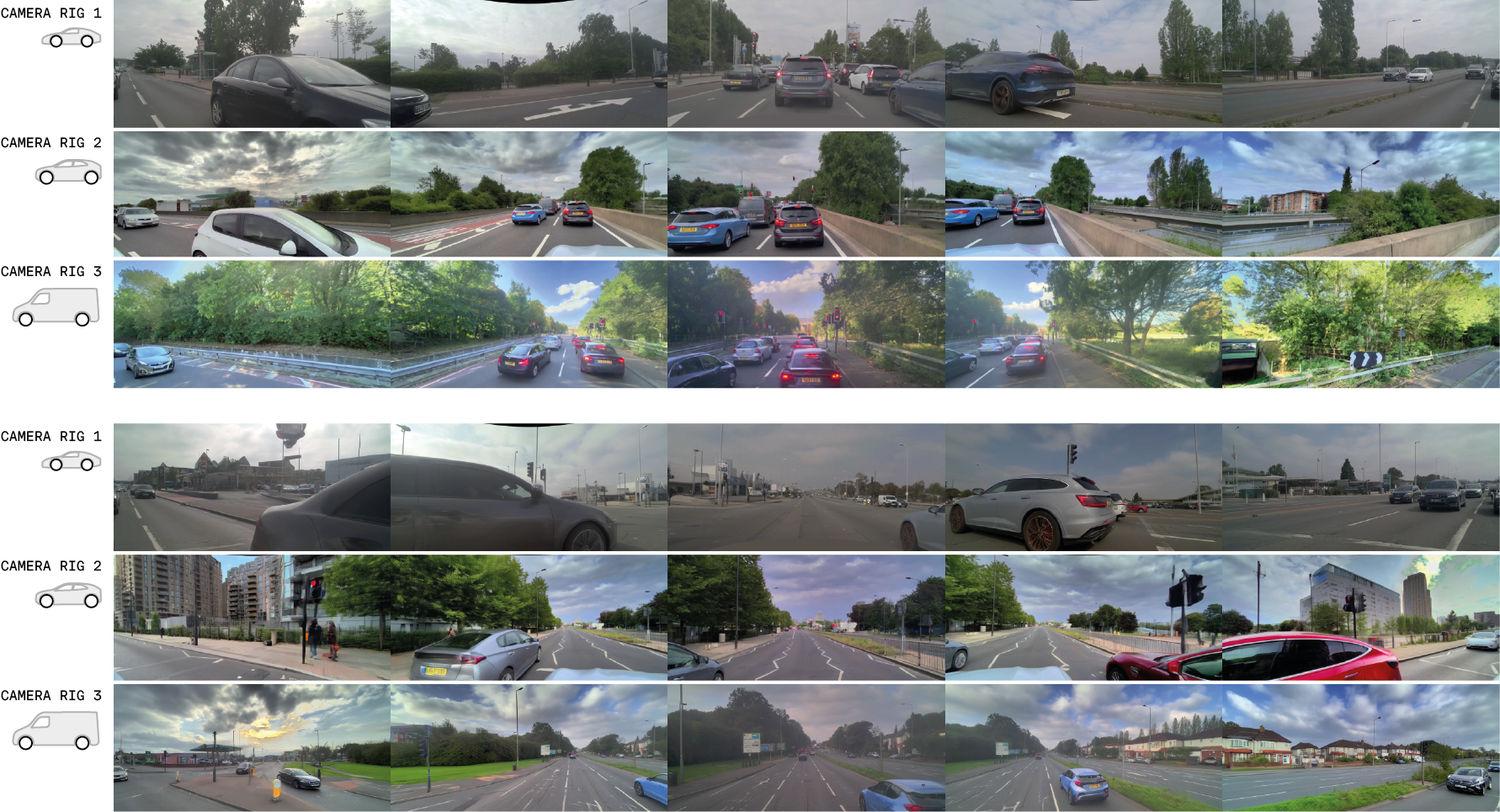}
    \caption{\textbf{Multi-rig video generation}. GAIA-2 supports various vehicle platforms and camera setups, maintaining spatial and temporal consistency. The two examples shown include camera arrangements from a sports car, an SUV, and a large van.}
    \label{fig:camera-rigs}
\end{figure}

\begin{figure}[t]
  \centering
  \makebox[\textwidth][c]{%
    \includegraphics[width=1\textwidth]{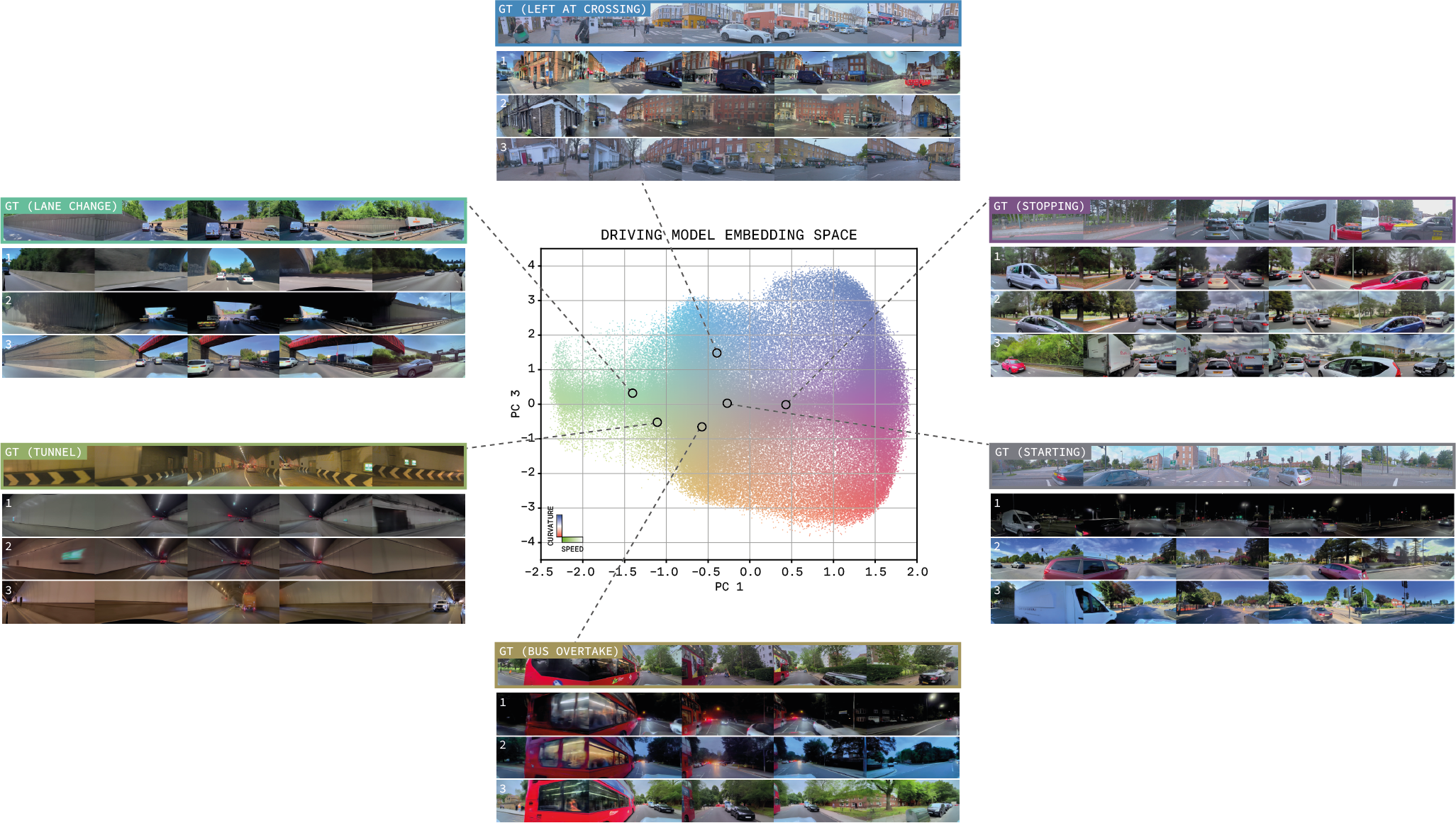}
    }
  \caption{\textbf{Scenario embedding conditioning}. Scenario embeddings from a proprietary driving model enable GAIA-2 to generate diverse yet semantically consistent variations from real-world scenarios. For each scenario type the top row shows the real data the embedding is derived from, and the bottom three rows show synthetic variants of it, guided by the embedding as conditioning. We also optionally apply additional conditioning such as country, weather, time of day, or camera parameters to control diversity.}
  \label{fig:embeddings}
\end{figure}

\section{Results}
\label{section:results}

We evaluate GAIA-2 through qualitative examples and quantitative metrics that highlight its generative fidelity, controllability, and suitability for synthetic data generation in autonomous driving. A comprehensive set of generated video examples is available at \url{https://wayve.ai/thinking/gaia-2}.

\paragraph{Augmenting Real-World Data.}
GAIA-2 enables dataset augmentation through a range of techniques that allow for visual and contextual diversification of real-world sequences.

\subsection{Qualitative Examples}

\paragraph{Diverse Scenario Generation.}
GAIA-2 supports the generation of highly diverse driving scenarios, spanning multiple countries, weather conditions, times of day, and road layouts. In addition to structured conditioning via metadata, GAIA-2 can be guided by CLIP embeddings, allowing control over scene semantics not explicitly captured in labels. This includes geographic and environmental context such as urban, mountainous, or coastal scenes (\Cref{fig:clip}).

Furthermore, due to its exposure to varied camera configurations during training, GAIA-2 can simulate driving videos across different vehicle embodiments. By conditioning on camera parameters, it maintains spatial and temporal consistency across viewpoints for rigs mounted on sports cars, SUVs, and large vans (\Cref{fig:camera-rigs}).

\begin{figure}[t]
  \centering
  \includegraphics[width=1.0\textwidth]{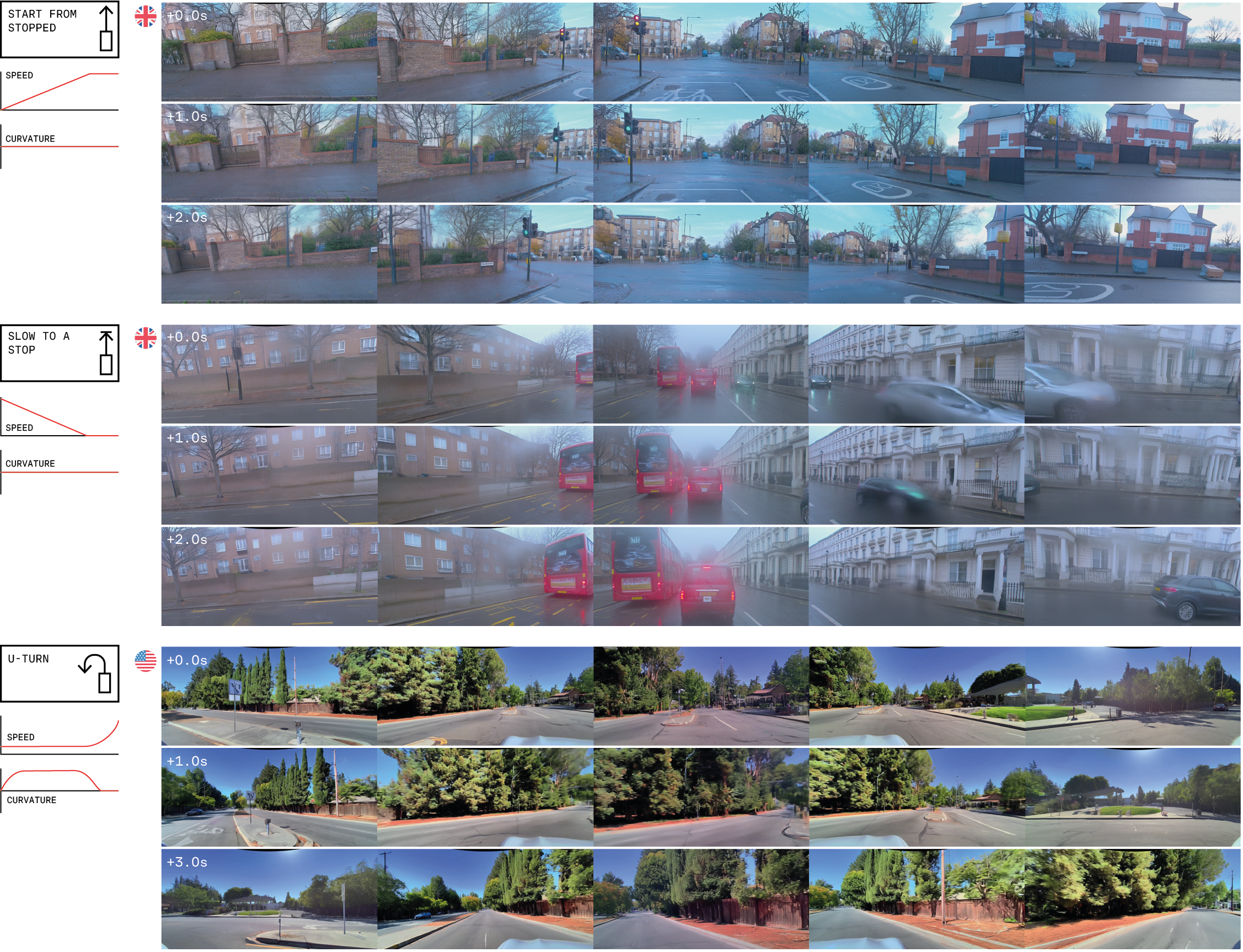}
  \caption{\textbf{Action-based generation}. GAIA-2 synthesizes diverse scenes from specified speed and curvature profiles. Each scenario is contextually plausible, despite the original video inputs being dropped out.}
  \label{fig:from-scratch-actions}
\end{figure}

\begin{itemize}
\item \textbf{Partial noise and denoise:} Latents derived from real videos are partially noised and then denoised under altered conditioning, such as different weather or lighting. This approach preserves semantics and ego motion while yielding diverse visual outputs (Figure~\ref{fig:noisedenoise}). We demonstrate the ability to modify environmental aspects of the scene while leaving core semantic and functional components as per the original. Essentially turning a single real-world example into multiple new scenarios: the same ego trajectory but visually diversified to take place during sunshine, rain, at sunset, at night, and in the snow.
\item \textbf{Scenario embeddings:} Scenario embeddings derived from our proprietary driving models give GAIA-2 the ability to generate multiple plausible variations from a single real-world example, providing many new examples of diverse yet contextually coherent synthetic data (\Cref{fig:embeddings}). Because the embedding space is meaningfully organized by the driving model, we are able to reliably generate scenarios that are semantically interpretable given their location in the embedding space, such as accelerating, decelerating, or overtaking other agents. By additionally conditioning on environmental factors, we can synthetically expand our dataset by increasing coverage across various conditions. By conditioning on different camera parameters, we can effectively replicate our corpus across any given vehicle platform.
\item \textbf{Action-based generation:} GAIA-2 can synthesize new scenes purely from ego-vehicle action trajectories. Given the speed and curvature profiles, it generates plausible visual contexts aligned with those dynamics, such as traffic light changes, braking behavior, or turning maneuvers (\Cref{fig:from-scratch-actions}). We show three examples: (1) conditioning on a `\textit{start from stopped}' speed profile, GAIA-2 generates plausible visual observations to fit those actions, in this case, a UK traffic light turning from red and amber to green; (2) conditioning on a `\textit{slow to a stop}' speed trajectory, GAIA-2 generates a scenario where the ego agent is slowing down behind a London taxi; and (3) conditioned on a strong leftward curvature and slow ramping speed, GAIA-2 generates a U-turn at a US intersection.
\end{itemize}

\begin{figure}[t]
  \centering
  \includegraphics[width=1.0\textwidth]{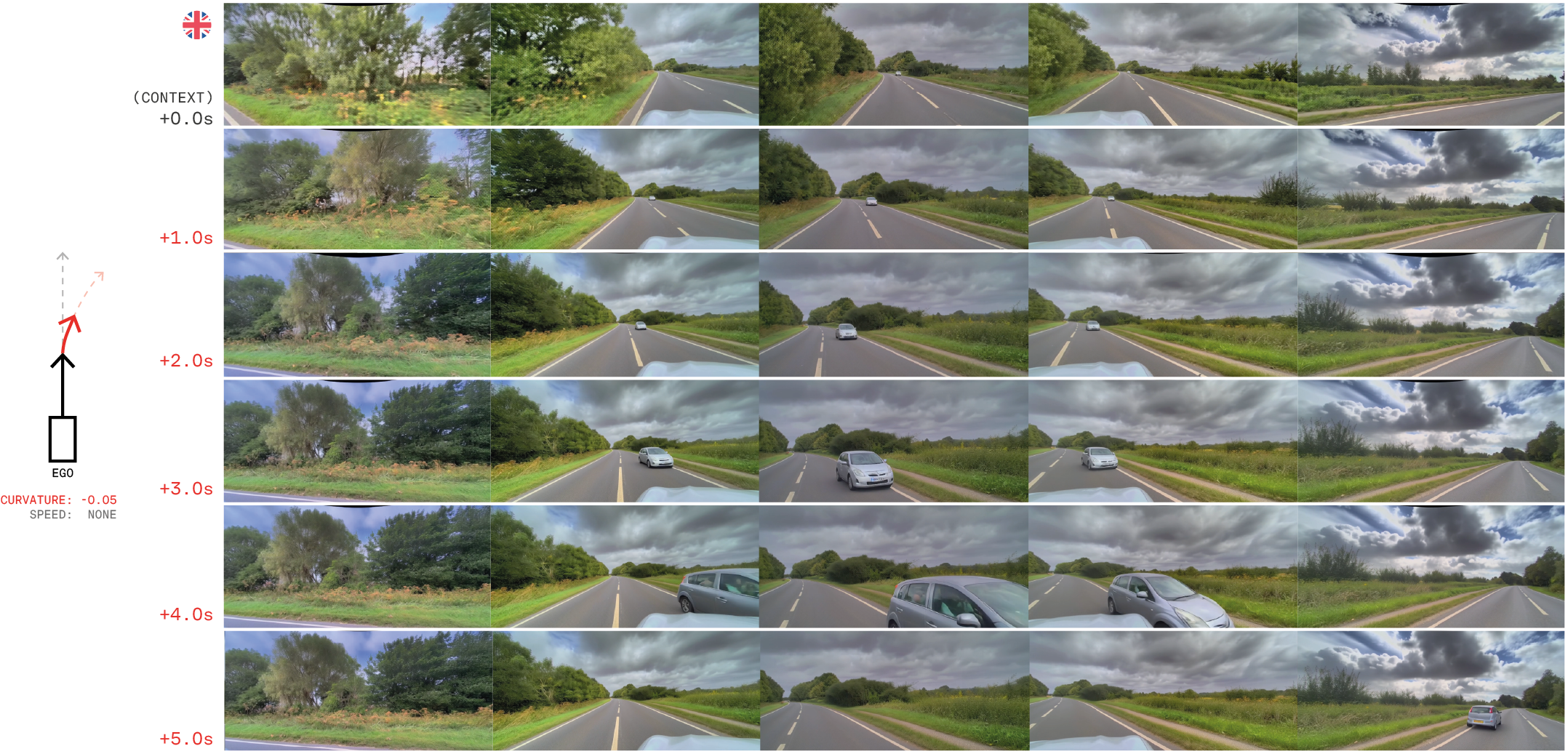}
  \caption{\textbf{Ego-induced safety-critical scenarios}. By applying unsafe action conditioning, GAIA-2 can generate hazardous situations, such as steering into oncoming traffic. We provide real video frames as context and GAIA-2 rolls out from these starting frames given some new action conditioning. In this example, we control the ego vehicle steering by conditioning on rightward curvature. We drop out speed so that GAIA-2 can generate multiple diverse outcomes. Note how the ego vehicle slows down while veering sharply into oncoming traffic. The oncoming vehicle swerves to avoid the ego vehicle.}
  \label{fig:unsafe-actions}
\end{figure}

\begin{figure}[t]
  \centering
  \makebox[\textwidth][c]{%
    \includegraphics[width=1.0\textwidth]{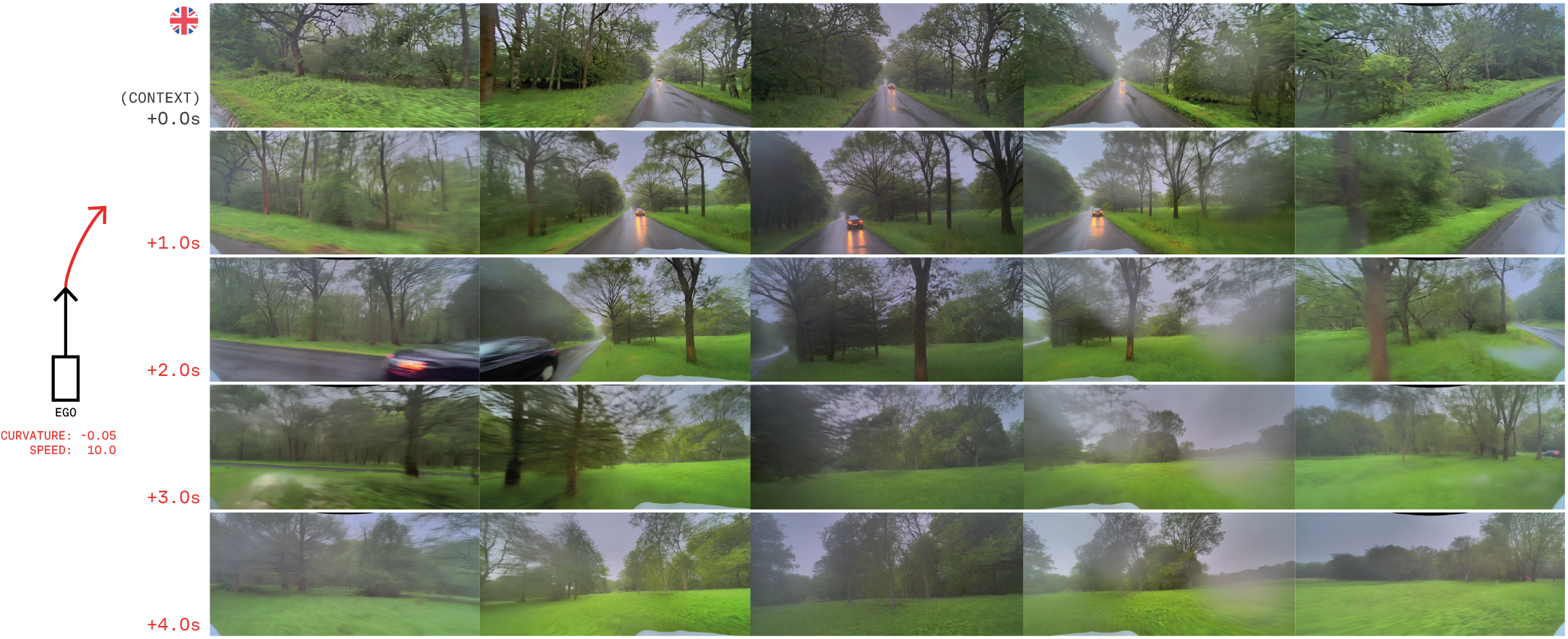}%
  }
    \caption{\textbf{Extreme generalization behavior}. When conditioned on high speed and strong curvature, GAIA-2 extrapolates off-road trajectories, capturing rare behaviors like driving into fields or forests.} 
  \label{fig:ood}
\end{figure}

\begin{figure}[t]
  \centering
  \includegraphics[width=1.0\textwidth]{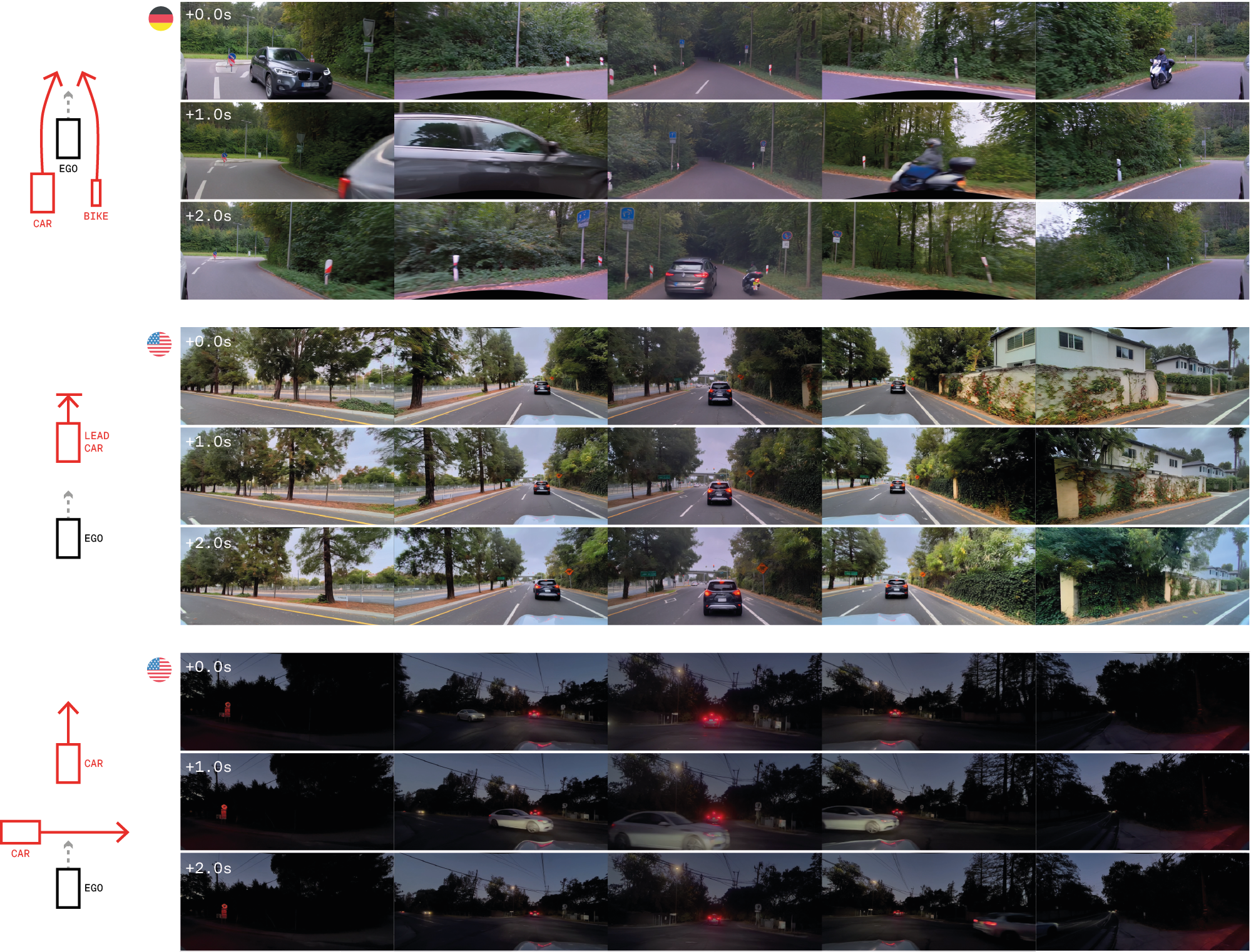}
  \caption{\textbf{Other-agent induced safety-critical scenarios}. 3D bounding box conditioning enables fine control over dynamic agents, allowing simulation of risky behaviors such as sudden braking, overtaking, or speeding through intersections.}
  \label{fig:cuboids}
\end{figure}

\paragraph{Generating Safety-Critical Scenarios.}
GAIA-2 is able to generate rare but safety-critical scenarios that would otherwise occur infrequently in the real world. We consider two classes of safety-critical situations: those caused by the ego agent, and those caused by other agents.
\begin{itemize}
\item \textbf{Ego-vehicle induced scenarios:} By conditioning on extreme or unsafe ego-vehicle actions (e.g., abrupt steering into oncoming traffic), GAIA-2 generates realistic scenarios critical to testing autonomous system resilience under hazardous conditions (\Cref{fig:unsafe-actions} and \Cref{fig:ood}).
\item \textbf{Other-agent induced scenarios:} Using 3D bounding boxes conditioning, GAIA-2 precisely controls the placement and motion of other agents, creating scenarios involving aggressive driving behaviors, emergency braking, or hazardous crossings, crucial for testing reactive capabilities of autonomous systems (\Cref{fig:cuboids}).
\end{itemize}

\begin{figure}[t]
  \centering
  \includegraphics[width=1.0\textwidth]{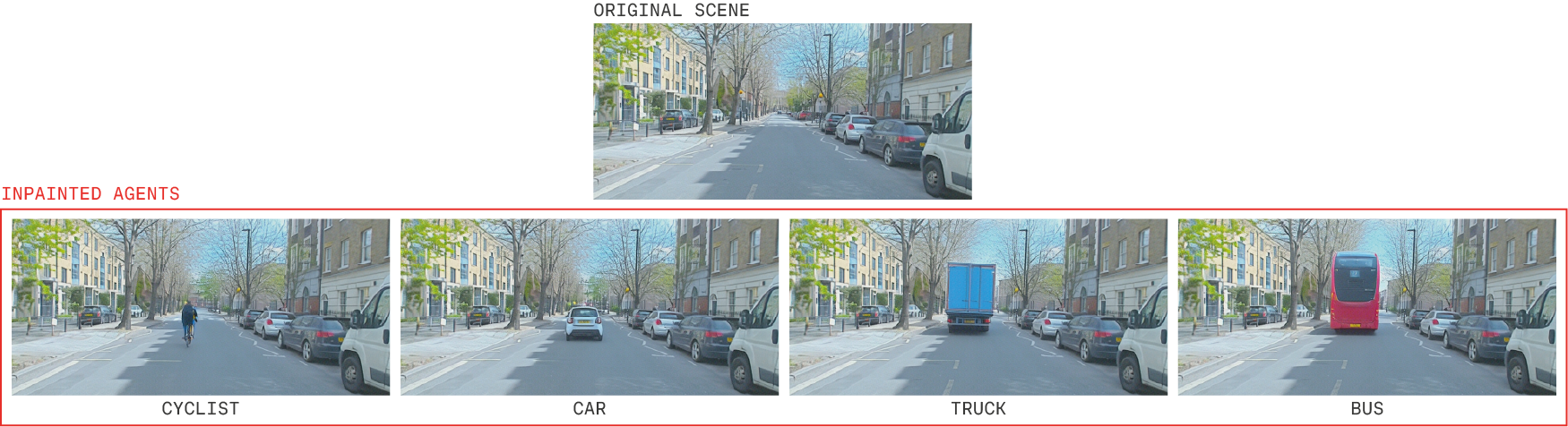}
  \caption{\textbf{Agent inpainting}. GAIA-2 can insert dynamic agents into masked regions of a video, guided by 3D bounding box conditioning. Background consistency and semantic realism are preserved.}
  \label{fig:inpainting}
\end{figure}

\paragraph{Inpainting.}
GAIA-2 supports spatial-temporal inpainting, enabling selective content editing. When provided with masked regions and corresponding agent-level conditioning, GAIA-2 inserts agents seamlessly into existing contexts without disrupting unrelated areas (\Cref{fig:inpainting}).

These qualitative results highlight GAIA-2’s ability to perform fine-grained and controllable video generation, supporting its use in scalable simulation and diverse scenario creation for autonomous systems.

\subsection{Metrics}
To quantitatively evaluate GAIA-2, we use a suite of metrics that assess visual fidelity, temporal consistency, and conditioning accuracy.

\paragraph{Visual Fidelity.}
Similarly to \cite{stein2023exposingflawsgenerativemodel}, we use the Frechet DINO Distance (FDD) and Frechet Inception Distance (FID) \cite{heusel2018ganstrainedtimescaleupdate} to measure the distance between the distributions of generated and real videos. The feature encoder is DINOv2 \cite{oquab2023dinov2} ViT-L/14, and the input images are bilinearly rescaled to a higher resolution $448 \times 952$ compared to $299 \times 299$ for InceptionV3 \cite{szegedy2015rethinkinginceptionarchitecturecomputer} in FID. We observe that FDD is less noisy than FID and saturates much later in training.

\paragraph{Temporal Consistency.} To evaluate the temporal consistency of the world model, we use the Frechet Video Motion Distance (FVMD) \cite{liu2024frechetvideomotiondistance}. This metric compares distributions of explicit key-point motion features for generated and ground-truth videos. We find this metric to be more aligned with human preferences for temporal consistency compared to FVD \cite{unterthiner18}.



\begin{figure}[t]
    \centering
    \includegraphics[width=1\textwidth]{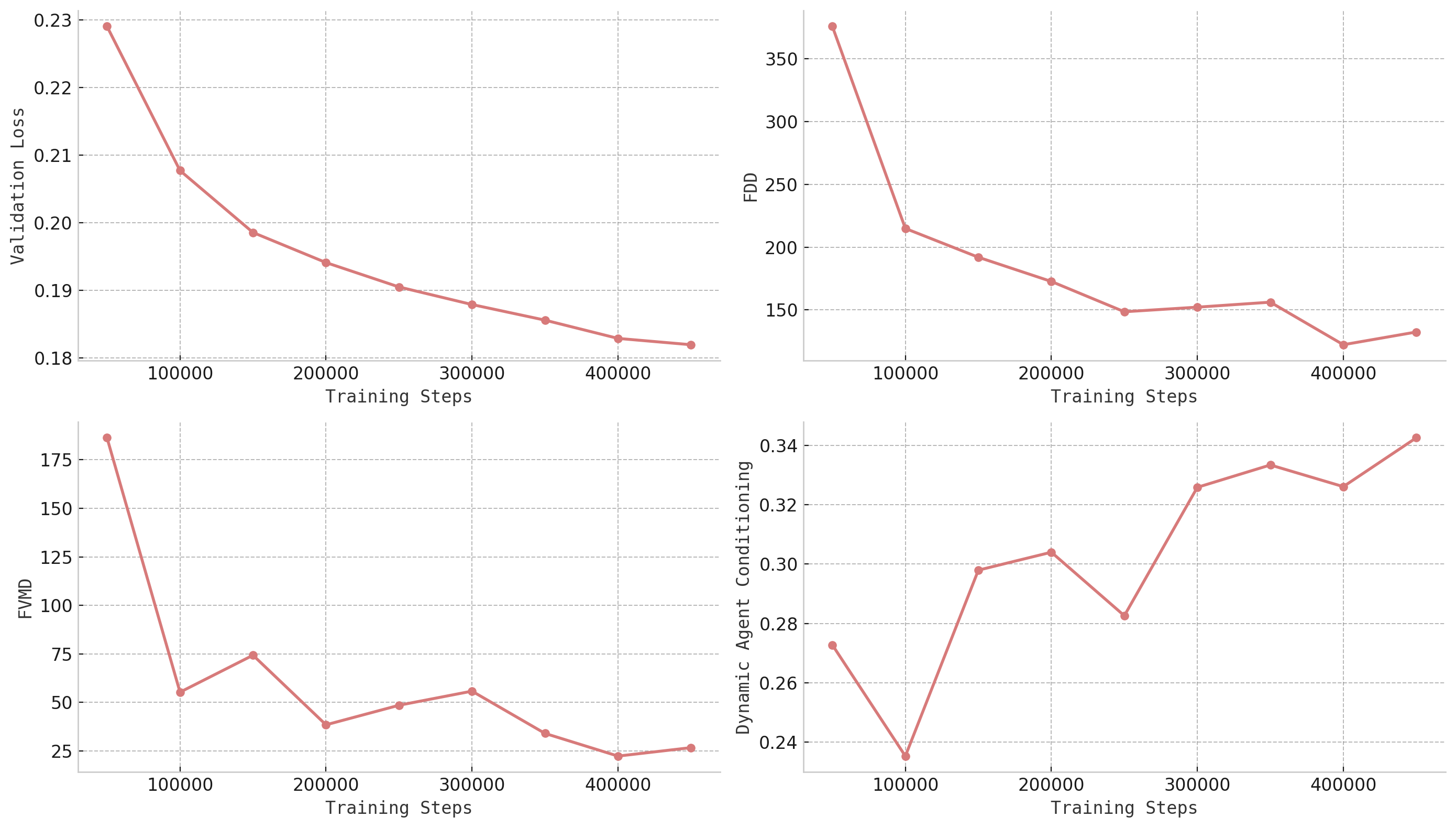}
    \caption{Validation loss and other metrics (evaluated on $n=1024$ samples). We found that the validation loss correlates well with human perceptual preferences.} 
    \label{fig:metrics}
\end{figure}

\paragraph{Dynamic Agent Conditioning.}  We evaluate dynamic agent fidelity by comparing the projection of the 3D bounding boxes (our conditional signal) with segmentation masks obtained via OneFormer \cite{jain2022oneformertransformerruleuniversal} on the generated video. We then compute class-based Intersection-over-Union (IoU) to measure how well the model adheres to the spatial and categorical specifications of dynamic agents during generation.



We report the metrics and the validation loss in \Cref{fig:metrics}. The validation loss and other metrics are evaluated on the generation from scratch task with $n=1024$ samples. Despite some noise from small sample size, all metrics exhibit a positive trend throughout training. Notably, the validation loss shows the strongest correlation with human preference, making it a valuable proxy for perceptual quality.


\section{Related Work}

\paragraph{Video Generative Models.}
Recent advancements in video generative models have mainly been driven by models that leverage latent representations derived through dimensionality reduction. These representations can be either continuous~\cite{kingma14,nvidia2024cosmos} or quantized~\cite{oord17,esser21}, and the success of video generation heavily depends on how efficiently these representations capture spatiotemporal information. One of the key challenges lies in ensuring that the mappings to and from latent space preserve both visual fidelity and semantic coherence. While scaling to large datasets improves generalization~\cite{agarwal2025cosmos}, challenges remain in preserving both visual realism and temporal consistency.

Autoregressive and masked generative models using quantized tokens have demonstrated strong modeling of temporal dependencies and motion dynamics~\cite{yan2021videogpt,moing21,ge2022long,seo22,hawthorne22,micheli2022transformers,teco22yan,villegas22phenaki,yu2023magvit}. However, their sequential generation processes lead to slow inference and error accumulation, which can limit their scalability for long sequences or complex multi-agent scenes. In contrast, latent diffusion models offer a more parallelizable alternative, using iterative denoising to generate full video sequences. These models, including Stable Video Diffusion~\cite{blattmann2023stable}, DIAMOND~\cite{alonso2024diffusionworldmodelingvisual}, Sora~\cite{openai2024sora}, MovieGen~\cite{polyak2024movie}, Gen-3~\cite{runway2024gen3}, and Cosmos~\cite{agarwal2025cosmos}, have achieved impressive quality on text-to-video and image-to-video tasks, especially in open-domain settings. 

However, most of these works prioritize visual aesthetics and domain-agnostic generation, offering limited support for structured control over scene elements or agent behavior. This makes them less suitable for applications (such as autonomous driving) that require precise, semantic, and multi-modal scene control across space and time. In this context, GAIA-2 aims to bridge the gap between high-fidelity generation and domain-specific controllability required for safety-critical systems.

\paragraph{Generative World Models in Autonomous Driving.}
Unlike general-purpose video generation, generative world models for autonomous driving must satisfy stricter requirements: multi-agent interaction, ego-motion control, environmental diversity, and multi-camera coherence. Early work in this space includes GAIA-1~\cite{gaia1-2023}, which introduced text and ego action conditioning in a discrete world model with a video diffusion decoder to improve temporal consistency. CommaVQ~\cite{comma2023vq} similarly leveraged a causal transformer on discrete tokens for ego-motion control. Although these methods took important steps toward controllability, they often targeted single-camera settings and partial controllability via action or text conditioning, limiting their applicability to comprehensive driving simulations.

Subsequent approaches adopted latent diffusion for higher fidelity generation and introduced more flexible control. DriveDreamer~\cite{wang2023drivedreamer} adopts a latent diffusion model conditioned on 3D bounding boxes, HD maps, and ego actions, and further incorporates an action decoder to predict future ego actions. Drive-WM~\cite{wang2024driving} extends this approach to a multi-camera setting (up to six cameras surrounding the ego vehicle) and introduces controllability over ego actions, dynamic agents, and environmental conditions, such as weather and lighting. UniMLVG~\cite{chen2024unimlvg} similarly enables multi-view generation conditioned on text, camera parameters, 3D bounding boxes, and HD maps, while MaskGWM~\cite{ni2025maskgwm} extends UniMLVG to support longer video generations. Vista~\cite{gao2024vista} employs latent diffusion for generating high-resolution, long-duration videos, and Delphi~\cite{ma2024unleashing} uses a data-driven approach to guide video generation toward failure cases in driving models, thereby improving the synthetic data's training utility. More recently, GEM~\cite{hassan2024gem} further generalizes scene control to include ego motion, object dynamics, and even human poses, suggesting broader applicability across domains such as human motion synthesis and drone navigation. A recent line of work has explored 4D geometry-aware generation. DriveDreamer4D~\cite{zhao2024drivedreamer4d} uses video generative models to synthesize videos along novel trajectories and combines real and generated footage to optimize a 4D Gaussian Splatting model for closed-loop simulation. Similarly, DreamDrive~\cite{mao2024dreamdrive} leverages generative priors to construct high-quality 4D scenes from in-the-wild driving data.

Despite these advancements, most of these solutions only address subsets of the requirements for realistic driving simulation. Some are limited to single-camera settings, others lack structured agent-level control, and few integrate multi-view generation and agent-level semantics into a single framework. Additionally, many do not support fine-grained editing tasks such as inpainting or targeted scene modification, essential for data augmentation and scenario variation in simulation loops.

Motivated by these limitations, \textbf{GAIA-2} is designed as a unified latent diffusion framework that brings together: 
\begin{itemize}
    \item multi-camera generation (up to five views); 
    \item structured and semantic conditioning over ego action, dynamic agents, and scene-level metadata; 
    \item flexible inference modes, including generation from scratch, inpainting, scene editing, and long-horizon rollouts; and 
    \item support for external latent spaces like CLIP and driving scenario embeddings.
\end{itemize} 
By combining continuous latent representations with large-scale training on a geographically and environmentally diverse dataset, GAIA-2 delivers both visual fidelity and scenario controllability. It fills a critical gap between general-purpose diffusion models and the domain-specific needs of autonomous driving, namely, the ability to simulate realistic, controllable, multi-view driving scenes for robust, scalable training and evaluation.

\section{Conclusions and Future Work}
\label{section:conclusion}
We presented GAIA-2, a domain-specialized latent diffusion model for video generation, designed to meet the diverse and exacting requirements of simulation in autonomous driving. GAIA-2 sets a new benchmark in generative world modeling by unifying multi-camera coherence, structured semantic conditioning, and fine-grained control within a scalable architecture. The model supports controllable generation conditioned on ego-vehicle dynamics, environmental features, road layout, and dynamic agents, while maintaining spatial and temporal consistency across up to five camera views.

The architecture combines a space-time factorized latent diffusion world model with a compact and semantically meaningful video tokenizer, enabling high-fidelity generation across diverse driving environments. GAIA-2 supports multiple inference modes—including generation from scratch, autoregressive rollouts, spatial inpainting, and real-scene editing—each empowering systematic exploration of the driving scene space. The model excels at producing typical as well as rare safety-critical scenarios, thus significantly enhancing data diversity, scenario coverage, and validation rigor in autonomous driving systems.

By scaling controllability and scene diversity in simulation, GAIA-2 bridges the gap between real-world data limitations and the increasing demands of training and evaluating robust AI-driving models. In doing so, it serves as a powerful tool for accelerating iteration cycles, improving generalization to out-of-distribution conditions, and stress-testing AI agents across complex, long-tail scenarios.

\paragraph{Future Work.}
While GAIA-2 represents a significant step forward, several research directions remain open and will guide future iterations of this work: (\textit{i}) Like all generative models, GAIA-2 occasionally produces temporal or semantic inconsistencies, particularly in long-horizon or complex scenarios. Improving the reliability and consistency of video generation through better failure detection, refinement models, or constraint-aware sampling is a key challenge.
(\textit{ii}) Although GAIA-2 enables parallelized generation, real-time or near-real-time video synthesis remains computationally intensive. Future work will explore model distillation, efficient transformer variants, and inference-time acceleration techniques to improve deployment feasibility in real-world pipelines. 
(\textit{iii}) Despite supporting a rich set of conditionings, new efforts will target richer agent behavior modeling, more nuanced environmental conditions, and open-ended natural language control. Further enrichment of the training dataset with rare and safety-critical events—especially those involving human agents, complex infrastructure, or social interactions—will push the boundaries of realistic simulation.

Together, these directions will ensure that GAIA-2 and its successors continue to advance the role of generative models as core infrastructure in the development of safe, robust, and generalizable autonomous systems.

\begin{ack}
\begin{small}
The authors would like to thank and acknowledge the many individuals whose invaluable contributions made this work possible: Alex Persin, Ana-Maria Marcu, Aniruddha Kembhavi, Benoit Hanotte, Evgenii Kashin, Francesca Iovu, Giacomo Gallino, Giulio D'Ippolito, Jeremy Plassmann, Lorenzo von Ritter, Mohamed Nabil, Nikhil Mohan, Oleh Chernov, Pragya Kale, Remi Tachet, Rudi Rankin, Sahana Vankatesh, Sarah Belghiti, Sofía Dudas, Tilly Pielichaty, Vassia Simaiaki, Victor Delépine, Vincent Micheli, Zak Murez.
\end{small}
\end{ack}


\bibliography{bibliography}

\end{document}